# Symbolic Probabilistic Inference with Continuous Variables

Kuo-Chu Chang and Robert Fung
Advanced Decision Systems
1500 Plymouth Street
Mountain View, California    94043-1230

## Abstract

Research on Symbolic Probabilistic Inference (SPI) [2, 3] has provided an algorithm for resolving general queries in Bayesian networks. SPI applies the concept of dependency-directed backward search to probabilistic inference, and is incremental with respect to both queries and observations. Unlike traditional Bayesian network inferencing algorithms, SPI algorithm is goal directed, performing only those calculations that are required to respond to queries. Research to date on SPI applies to Bayesian networks with discrete-valued variables and does not address variables with continuous values.

In this paper[1], we extend the SPI algorithm to handle Bayesian networks made up of continuous variables where the relationships between the variables are restricted to be "linear gaussian". We call this variation of the SPI algorithm, SPI Continuous (SPIC). SPIC modifies the three basic SPI operations: multiplication, summation, and substitution. However, SPIC retains the framework of the SPI algorithm, namely building the search tree and recursive query mechanism and therefore retains the goal-directed and incrementality features of SPI.

## 1 Introduction

The Bayesian networks technology provides a language for representing uncertain beliefs and inference algorithms for drawing sound conclusions from such representations. A Bayesian network is a directed, acyclic graph in which the nodes represent random variables, and the arcs between the nodes represent possible probabilistic dependence between the variables. The success of the technology is in part due to the development of efficient probabilistic inference algorithms [5, 6, 7]. These algorithms have for the most part been designed to efficiently compute the posterior probability of each node or the result of simple arbitrary queries. They have not efficiently addressed the more general problem of answering multiple queries with respect to differing sets of evidence.

Recent research in Symbolic Probabilistic Inference (SPI) [2, 3] has made a significant step in this direction. Unlike traditional Bayesian network inference algorithms, SPI algorithm is goal directed, performing only those calculations that are required to respond to queries. In addition, SPI is incremental with respect to both queries and observations. However, the research to date on SPI applies only to Bayesian networks with discrete-valued variables.

There have been several inference algorithms designed to handle networks that are made up of continuous variables where the relationships between the variables are restricted to be "linear gaussian". The algorithms include the distributed algorithm [7] and the influence diagram approach [4, 8].

In this paper, we extend the SPI algorithm to perform this function—handle Bayesian networks with continuous linear gaussian variables. We call the extension, SPI Continuous (SPIC). The framework of this algorithm is the same as that for SPI. However, the basic SPI operations of multiplication, integration and substitution are quite different. Because SPIC stays within the SPI framework, the goal-directed and incrementality features of the algorithm are preserved.

The paper is organized as follows. Section 2 briefly summarize the SPI algorithm which includes the construction of the SPI node tree and the recursive query processing. Section 3 describes the new algorithm with continuous variables. The representation are described as well as the "basic" operations. Finally, some concluding remarks are given in Section 4.

---

[1] This work is based on research supported by WRDC under Contract F33615-90-C-1482.

## 2 Overview of the SPI Algorithm

The SPI algorithm consists of several major processing steps. The first step is to organize the nodes of a Bayesian network into a tree structure for query processing. We call these structures SPI trees. In the second step, queries are directed to the root node of the SPI tree. The query is decomposed into queries for the node's subtrees. This recursive procedure continues until a particular query can be answered at the node at which it is directed. The answer is then computed and returned to the next higher level in the SPI tree. Once a node has responses from all of its subtrees it can compute its own response to its predecessor node. This process terminates when the root node processes all the responses from its subtrees.

An SPI tree is constructed by organizing the nodes of a Bayesian network into a tree structure. The only constraint on the construction process is that if there is an arc between two nodes in the original network, then one of the nodes must be a direct or indirect predecessor of the other in the SPI tree. This constraint allows many possible SPI trees.

The first step in building a SPI tree is to choose the root node. This done by computing the maximum node to node distance for each node. The node that has the smallest maximum distance is chosen as the root node. This heuristic is designed to produce a "bushy" SPI tree which can take advantage of distributed processing. The second step is to use maximum cardinality search [9] to build the tree from the root node. This step constructs the tree based on the connectivity principle and guarantees satisfaction of the tree construction constraint.

The general format for a query received by SPI is as a conditional probability, namely, $P\{X|Y\}$, where $X$ and $Y$ are sub sets of nodes in the Bayesian networks. To be processed by SPI, queries of this form are first transformed into another format. The format consists of two set of nodes $L$ and $M$ which satisfy:

$$M = (X \cup Y) \cap D(L) \qquad (1)$$

where $X = M \cap L, Y = M \setminus L$, and $D(L)$ is the dimension of the distribution associated with the node set $L$. Intuitively, $L$ represents the minimum set of node distributions needed to respond to the query and $M$ is dimension of the desired response. $L$ and $M$ can be computed in linear time and are simple to implement [3]. Figure 1 and 2 show a simple Bayesian network and the corresponding SPI tree. For example, if the query is to find the joint probability of $a_1$ and $c_2$, the query being sent to the root node will be consisting of $L = \{a_1, a_2, c_1, c_2\}$ and $M = \{a_1, c_2\}$.

The heart of the SPI algorithm is as follows; at any node $i$, a request arrives for a probability distribution represented by $L$ and $M$, the algorithm responds to the request by computing the "generalized" distribu-tion $Q(M)$ [3]. $Q(M)$ is obtained by multiplying the distributions in node $L$ and summing over dimensions $L \setminus M$. If such a distribution had already been computed earlier and cached, it can be returned immediately. However, usually it will be necessary to send requests to node's successors in the SPI tree in order to compute the response. It is obvious that if a particular subtree has nothing to do with the query (i.e., there is no intersection), then no query will be sent to that subtree. For the same example above, the query can be obtained as,

$$Q(a_1, c_2) = \pi_{a_1} \sum_{a_2, c_1} \pi_{a_2} \pi_{c_1} \pi_{c_2} \qquad (2)$$

where $\pi_i$ represents the probability distribution associated with node $i$. There are three major operations in SPI algorithm: multiplication, summation and substitution. Multiplication calculates the product of two distributions; summation calculates the sum of a distribution over a set of variables; and substitution calculates the result of substituting an observed value for a node into a distribution. For networks with continuous variables, the SPI algorithm can be applied directly. However the multiplication, summation, and substitution operations must be modified. In the next section, we will describe the corresponding multiplication, integration, and substitution operations for the networks with continuous variables.

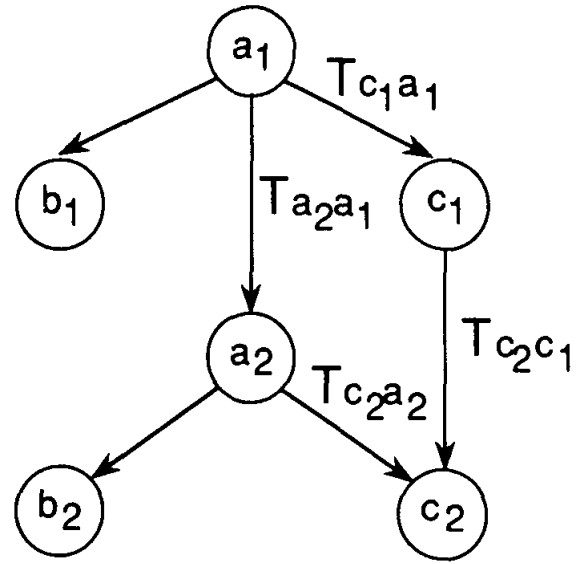

Figure 1: An Example Network

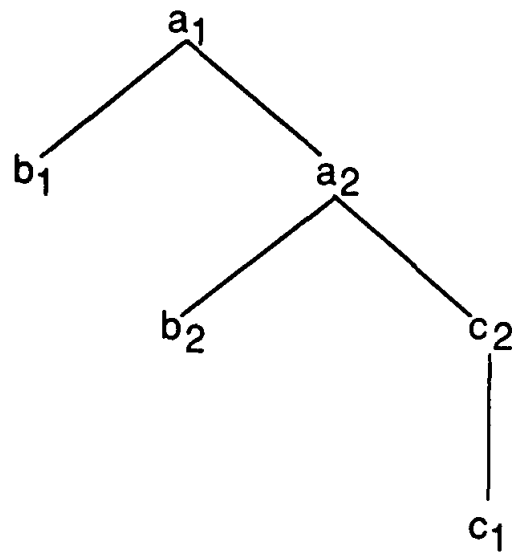

Figure 2: The SPI tree

## 3 The SPI with Continuous Variables Algorithm

The continuous SPI (SPIC) algorithm requires redefinition of the SPI operations: multiplication, integration, and substitution. The general mechanism for the continuous SPI algorithm is basically the same as the discrete one except for the caching operation and the handling of evidence. We first describe the representation for conditional probability distributions in linear gaussian continuous variables and then the three operations in detail.

### 3.1 Node Representation

A SPIC node represents a vector of continuous variables. SPIC restricts the conditional probability distribution of each node to be "linear gaussian". "Linear gaussian" distributions are the sum of a deterministic component and a probabilistic component. The deterministic component is a linear combination of the node's predecessor values. The probabilistic component is restricted to be gaussian (i.e., normal) which can be specified with mean vectors and covariance matrices. For a Bayesian network of this type, the necessary prior information needed before any inference can be drawn are the the prior distributions of the root nodes (i.e., mean and covariance) and the links between nodes in the network.

For example, for a random vector represented by a node $x$, if it is a root node, only mean $\bar{x}$ and the corresponding covariance matrix $Q_x$ need to be specified. If it is not a root node, and has a set of predecessor nodes $x_1^p, .., x_n^p$, then the relation between $x$ and its predecessors represented by the following linear equation need to be specified,

$$x = B_1 x_1^p + ... + B_N x_N^p + w_x \qquad (3)$$

where $B_1, .., B_N$ are constant transition matrices representing the relative contribution made by each of the predecessor variables to the determination of the dependent variable $x$, and $w_x$ is a noise vector summarizing other factors affecting $x$. $w_x$ is assumed to be normally distributed with mean $\bar{x}$ and covariance $Q_x$. For most applications, $w_x$ will have a zero mean. However this is not always true *apriori* and can occur when distributions are multiplied together (e.g., a root node and a non-root node).

For each node $x$ in a network, the sufficient information describing the node itself and the relationship to its predecessors can therefore be represented in the following form:

$$\{\bar{x}, Q_x, (B_1, x_1^p)..(B_N, x_N^p)\} \qquad (4)$$

With this simple representation, we will then describe the multiplication, integration, and substitution operations.

### 3.2 Multiplication

In this section, we describe the multiplication operation for SPIC. We describe when distributions can be multiplied, what the result will look like, and then describe in detail how each part of the result is computed.

There is a constraint on what distributions can be multiplied. This constraint called "combinability" was developed in [1]. According to the theorem derived in [1], a set of nodes $S$ is "combinable" (i.e., able to be aggregated) if and only if every pair of nodes in the set is combinable. Two nodes are combinable if all nodes in the path(s) between the two nodes are in the set $S$. It can be shown that if the set of nodes corresponding to the distributions to be multiplied are "combinable", then multiplying the distributions is the same as finding the "joint" distribution of those nodes, or in other words, to find the new probability representation of the "combined" node.

Based on the separation principle in the SPI tree (i.e., any node separate its successors rooted from itself) [3], it can be easily shown that any distributions that will be multiplied from any query request during SPI processing are always combinable. In other words, we can transform the multiplication operation in SPI algorithm into a node combination operation for the continuous nodes.

80  Chang and Fung

The dimension of the resulting distribution will be the sum of the dimensions of those nodes to be multiplied. For instance, two representations of nodes (variables) $x_1$ and $x_2$ each with dimension $D(x_1)$ and $D(x_2)$ are to be multiplied, the resulting representation will have dimension $D(x_1) + D(x_2)$, which can be interpreted as the description of the combined variable $x$ with $x_1$ stack over $x_2$, i.e., $x = \begin{bmatrix} x_1 \\ x_2 \end{bmatrix}$. The question now is how to calculate the new probability representation of $x$ based on the old ones of $x_1$ and $x_2$. It is clear that this is nothing but to identify the new predecessors $x_i^P$ of $x$ and calculate the links $B_i$ between $x$ and the new predecessors as well as to calculate the new conditional mean $\bar{x}$ and the associated covariance $Q_x$.

First of all, the new predecessors of $x$ is just the union of the predecessors of $x_1$ and $x_2$ (excluding $x_1$ or $x_2$ when one is the direct predecessor of the other). The new linear relation (coefficients) between $x$ and $x_i^P$ are obtained based on the old links. Conceptually, this can be accomplished by first breaking down the combined node into the original element nodes, and then find the relation between the predecessor and the element nodes individually. To do so, all paths between the predecessor and the desired element node are found and then combined together. The contribution of a path is obtained by multiplying the transition matrices of links along the path in the original network.

The mean and associated noise covariance matrix for the combined node is the last part of the representation that needs to be calculated. The first quantity that requires computation is the implicit linear relationship $T$ between the two nodes that are to be multiplied. Based on the SPI tree structure, it can be shown that the $T$ for two nodes is non-zero only when one node is the direct or indirect predecessor of the other. If the nodes do stand in this relation, $T$ can be calculated by first finding all paths between the two nodes and adding the contribution from all paths. As in the process for finding the link coefficients for the new predecessors, the path contributions are obtained by multiplying together the transition matrices of links along the path. Given $T$, the new mean and covariance matrix of $x$ can be obtained as below, given that $x_1$ is the direct or indirect predecessor of $x_2$.

$$\bar{x} = \begin{bmatrix} \bar{x}_1 \\ T\bar{x}_1 + \bar{x}_2 \end{bmatrix} \quad (5)$$

$$Q_x = \begin{bmatrix} Q_{x_1} & Q'_{x_1}T' \\ TQ_{x_1} & TQ_{x_1}T' + Q_{x_2} \end{bmatrix} \quad (6)$$

For example, the results of multiplying $\pi_{c_1}$ and $\pi_{c_2}$ of the previous example given in Figure 1 can be represented by

$$c_{12} = \begin{bmatrix} c_1 \\ c_2 \end{bmatrix} = \begin{bmatrix} T_{c_1 a_1} \\ T_{c_2 c_1} T_{c_1 a_1} \end{bmatrix} a_1 + \begin{bmatrix} 0 \\ T_{c_2 a_2} \end{bmatrix} a_2 + \begin{bmatrix} W_{c_1} \\ T_{c_2 c_1} W_{c_1} + W_{c_2} \end{bmatrix} \quad (7)$$

where $c_{12}$ is the "combined node", $T_{xy}$ are the links (transition matrices) between $x$ and $y$, and $a_1$ and $a_2$ are the new predecessors. In addition to the links obtained as above, the new conditional mean and corresponding covariance are obtained as,

$$\bar{c}_{12} = \begin{bmatrix} \bar{c}_1 \\ \bar{c}_2 + T_{c_2 c_1} \bar{c}_1 \end{bmatrix} \quad (8)$$

which is zero since both $\bar{c}_1$ and $\bar{c}_2$ are zeros, and

$$Q_{c_{12}} = \begin{bmatrix} Q_{c_1} & Q'_{c_1} T'_{c_2 c_1} \\ T_{c_2 c_1} Q_{c_1} & T_{c_2 c_1} Q_{c_1} T'_{c_2 c_1} + Q_{c_2} \end{bmatrix} \quad (9)$$

In the discrete case, multiplying distributions can be expensive since the size of the resulting distribution grows exponentially with the number of distribution (nodes) to be multiplied. However, in the linear gaussian case, the resulting representation only increases quadratically. This is because a gaussian distribution can be sufficiently represented by a mean vector and a covariance matrix.

### 3.3 Integration

The integration operation is relatively simple. All one has to do is to identify and keep the appropriate slots from the representation (i.e., links to predecessors, mean, and covariance) and discard the rest of them. It can be easily shown that the reduced representation precisely describes the resulting distribution after the integration. For example, for the distribution of the combined node $c_{12}$ obtained from the multiplication above. If the goal is to integrate $c_1$ out of the joint probability representation, all one has to do is to grab the appropriate slots from the links to $a_1$ and $a_2$, the mean vector $\bar{c}_{12}$, and the covariance matrix $Q_{c_{12}}$. These slots are corresponding to the $c_2$ variable, namely,

$$c_2 = T_{c_2 c_1} T_{c_1 a_1} a_1 + T_{c_2 a_2} a_2 + T_{c_2 c_1} W_{c_1} + W_{c_2} \quad (10)$$

$$\hat{c}_2 = \bar{c}_2 + T_{c_2 c_1} \bar{c}_1 \quad (11)$$

and

$$\hat{Q}_{c_2} = T_{c_2 c_1} Q_{c_1} T'_{c_2 c_1} + Q_{c_2} \quad (12)$$

### 3.4 Substitution

Evidence is represented in the form of exact observation of the values of variables. The set of variables which have been observed is denoted by $E$. One easy way of incorporating evidence is to include evidence in the query, for instance, $P\{X|Y,E\}$ and then substitute the observed values $E^*$ for $E$ after the more general query is computed. Suppose the query results before the substitution of the observation is represented as

$$X = \bar{X} + K_Y Y + K_E E \tag{13}$$

with the associated covariance matrix $\Sigma_X$. To "substitute" the observation $E^*$, we first remove the link $K_E$ from the representation, then replace the mean $\bar{X}$ by $\bar{X} + K_E E^*$. The covariance matrix remains the same. It can be easily shown that the new representation correctly describe the results of the query, $P\{X|Y, E = E^*\}$. Other more efficient methods such as to do substitution before query are currently under our investigation.

## 4  Conclusion

Recent research in Symbolic Probabilistic Inference (SPI) has made a significant step in improving efficiency of general query processing. Unlike traditional inference algorithms, SPI algorithm is goal directed, performing only those calculations that are required to respond to queries. In addition, SPI is incremental with respect to both queries and observations. In this paper, we extent the SPI algorithm to handle Bayesian networks with linear gaussian variables. We call the algorithm SPIC. The framework of this algorithm is the same as that for SPI. However, the basic operations of multiplication, integration and handling observation are quite different. The equivalent operation to the multiplication of discrete case is similar to the "node combination" operation in the continuous case and the equivalent operation to the summation of discrete case is the integration operation. Because SPIC stays within the SPI framework, the goal-directed and incrementality features of the algorithm are preserved.

In this paper, we have only addressed the problem of continuous variables that are restricted to have linear gaussian models. Many real-world problems may require nonlinear or nongaussian models. Classical methods such as approximation with linearization (e.g., extended Kalman filters) or sum of gaussians deserve attention in further investigation. A version of the SPIC algorithm has been implemented, preliminary results show expected performance.